\documentclass[conference,letterpaper]{IEEEtran}

\usepackage[utf8]{inputenc} 
\usepackage[T1]{fontenc}
\usepackage{url}
\usepackage{ifthen}
\usepackage{array}
\usepackage{multirow}
\usepackage{multicol}
\usepackage{makecell}
\usepackage{tabularx}
\usepackage{caption}
\usepackage{booktabs}
\usepackage{bm}
\usepackage{dsfont}
\usepackage{cite}
\usepackage{float}
\usepackage{hyperref}
\usepackage{algorithm2e}
 \usepackage[cmex10]{amsmath} 
\usepackage{amsthm}
\usepackage{amsmath,amssymb,amsthm}
\usepackage{algorithm2e}
\usepackage{graphicx}
\newcommand{\mbhY}{\widehat{\mathbf{Y}}^{u}}

\newcommand{\mbXl}{\mathbf{X}_n^l}
\newcommand{\mbXu}{\mathbf{X}_m^u}
\newcommand{\mbXlu}{\mathbf{X}^{l,u}}
\newcommand{\mbZl}{\mathbf{Z}^{l}}
\newcommand{\mbZup}{\hat{\mathbf{Z}}}

\newcommand{\mbYl}{\mathbf{Y}^{l}}
\newcommand{\mbY}{\mathbf{Y}}
\newcommand{\unif}{\mathrm{Unif}(k)}

\newtheorem{theorem}{Theorem}
\newtheorem{proposition}{Proposition}

\newtheorem{corollary}{Corollary}
\theoremstyle{definition}

\newtheorem{assumption}{Assumption}
\theoremstyle{remark}
\newtheorem{remark}{Remark}

\begin{document}
\title{Robust Semi-supervised Learning via $f$-Divergence and $\alpha$-R\'enyi Divergence} 



\author{%
  \IEEEauthorblockN{Gholamali~Aminian\IEEEauthorrefmark{1},
                    Amirhossien~Bagheri\IEEEauthorrefmark{2},
                    Mahyar~JafariNodeh\IEEEauthorrefmark{3}\IEEEauthorrefmark{4},
                    Radmehr~Karimian\IEEEauthorrefmark{2},
                    Mohammad-Hossein~Yassaee\IEEEauthorrefmark{2}}
  \IEEEauthorblockA{\IEEEauthorrefmark{1}%
                   The Alan Turing Institute, British Library, 96 Euston Rd., London, UK, gaminian@turing.ac.uk}
    \IEEEauthorblockA{\IEEEauthorrefmark{3}%
    Sharif University of Technology, Iran,
   \{amir.bagheri, radmehr.karimian, yassaee\}@sharif.edu}
  \IEEEauthorblockA{\IEEEauthorrefmark{3}%
                    Institute for Data, Systems and Society (IDSS), Massachusetts Institute of Technology, USA,
                    mahyarjn@mit.edu}
 \IEEEauthorblockA{\IEEEauthorrefmark{4}%
Laboratory for Information and Decision Systems (LIDS), Massachusetts Institute of Technology, USA}                
}



\maketitle

\begingroup\renewcommand\thefootnote{$^*$}
\footnotetext{ The corresponding author is the first author. The authors' names are listed in alphabetical order.}
\endgroup


\begin{abstract}
    This paper investigates a range of empirical risk
    functions and regularization methods suitable for self-training
    methods in semi-supervised learning. These approaches draw
    inspiration from various divergence measures, such as $f$-
    divergences and $\alpha$-R\'enyi divergences. Inspired by the theoretical
    foundations rooted in divergences, i.e., $f$-divergences and $\alpha$-R\'enyi
    divergence, we also provide valuable insights to enhance the
    understanding of our empirical risk functions and regularization
    techniques. In the pseudo-labeling and entropy minimization
    techniques as self-training methods for effective semi-supervised
    learning, the self-training process has some inherent mismatch
    between the true label and pseudo-label (noisy pseudo-labels)
    and some of our empirical risk functions are robust, concerning
    noisy pseudo-labels. Under some conditions, our empirical risk
    functions demonstrate better performance when compared to
    traditional self-training methods.
\end{abstract}
\vspace{-0.5em}
\section{Introduction}
Machine learning applications such as finance, natural language processing, and computer vision have access to vast amounts of data, but sometimes this data lacks labels. This lack of labeling poses a challenge to traditional supervised learning methods. Semi-supervised learning (SSL) techniques leverage both labeled and unlabeled data samples to improve performance in supervised learning scenarios. One such SSL technique is self-training algorithms, which are explored in~\cite{ouali2020overview}. These algorithms use confident predictions from a supervised model to assign labels to unlabeled data. There are two primary approaches to self-training-based SSL: entropy minimization and pseudo-labeling.

Entropy minimization methods use an entropy function as a regularization term, aiming to penalize uncertainty in label predictions for unlabeled data~\cite{grandvalet2005semi}. The underlying assumption behind entropy minimization algorithms can be attributed to either the manifold assumption~\cite{iscen2019label}, which assumes that labeled and unlabeled data samples are drawn from a standard data manifold, or the cluster assumption~\cite{chapelle2003cluster}, which suggests that similar data features tend to share the same label.

Pseudo-labeling, introduced in~\cite{lee2013pseudo}, involves training a model using labeled data and assigning pseudo-labels to the unlabeled data based on the model's predictions. These pseudo-labels are then used to construct another model, which is trained in a supervised manner using both labeled and pseudo-labeled data. However, neural network predictions may exhibit inaccuracies, particularly in neural networks. This issue is further exacerbated when these erroneous predictions are employed as labels for unlabeled samples, a characteristic inherent in the practice of pseudo-labeling. The phenomenon of overfitting to incorrect pseudo-labels generated by the neural network is widely recognized as confirmation bias~\cite{arazo2020pseudolabeling}.

This work proposes new empirical risk functions and regularizers based on the divergence between the empirical distribution data samples and conditional discrete distribution over the label set. These empirical risk functions are then applied to self-training approaches, i.e., pseudo-labeling and entropy minimization, in SSL applications. Our empirical risk functions are more robust to noisy pseudo-labels (i.e., the pseudo-label is different from the true label) of unlabeled data samples, which are generated by self-training approaches. Inspired by some divergences properties, we also provide an upper bound on the true risk of some empirical risk functions.

Our main contributions to this paper are as follows:
\begin{itemize}
    \item We propose novel risk functions inspired by different divergences, including $f$-divergences and $\alpha$-R\'enyi divergence.
    \item We combine our risk functions with self-training methods, i.e., pseudo-labeling and entropy minimization. For this purpose, we propose novel regularization terms inspired by $f$-divergences and $\alpha$-R\'enyi divergence.
    \item For some divergences, which are also metric distance, we provide an upper bound on ideal performance (access to all true labels for all unlabeled data) of the empirical and true risk functions.
    \item  We provide an empirical analysis of our empirical risk functions and regularizers under different scenarios and datasets to show their performance under noisy pseudo-labels.
\end{itemize}
\vspace{-0.5em}
\section{Preliminaries}
\subsection{Problem Formulation}
Throughout the paper, upper-case letters denote random variables (e.g., $Z$), lower-case letters denote the realizations of random variables (e.g., $z$), and calligraphic letters denote sets (e.g., $\mathcal{Z}$). 
All the logarithms are natural, and all the information measure units are nats. We denote the set of integers from 1 to $N$ by $ [N] \triangleq \{1,\dots,N\}$.
\newline
We denote the space of labels and features by $\mathcal{Y}$ and $\mathcal{X}$, respectively. The set of labeled and unlabeled data samples\footnote{We use features and data samples terms interchangeably.} are defined with  $\mbXl:=\{X_i^l\}_{i=1}^n$ and $\mbXu:=\{X_j^u\}_{j=1}^m$, where the $X_i^l$ and $X_j^u$ are the labeled and unlabeled data samples drawn of distribution $P_X$. The set of all labeled and unlabeled data samples is defined by $\mbXlu:=\mbXl \cup \mbXu$. The labeled dataset is denoted by $\mathbf{Z}_n^l$, which contains $n$ samples, $\mathbf{Z}_n^l=\{(X_i^l,Y_i^l)\}_{i=1}^n$, where $X_i^l\in\mathcal{X}^l\subset \mathcal{X}$ and $Y_i^l\in\mathcal{Y}$ are labeled features and the corresponding labels, respectively. For classification problems with $k$ classes, we consider $|\mathcal{Y}|=k$. We define the uniform distribution over $\mathcal{Y}$ with $\mathrm{Unif}(k)$. Let $\hat{P}(\mbY|X_i)$ denote the empirical distribution over labels given the feature $X_i$. Our model is able to predict the underlying conditional distributions of labels given features, i.e., $P_{\theta}(\mbY|X_i):=\{P_\theta(Y=j|X_i)\}_{j=1}^k$, where $\theta \in \Theta$ is the parameter of our model. This means that our model can estimate the probability of each possible label for each given feature vector. For example, the output of the Softmax layer in neural networks can be considered as an estimation of the conditional distribution of labels given the feature. We examine the following scenarios,
\begin{itemize}
    \item Supervised Learning (SL): We train the model based on only labeled data samples,
    \item Semi-Supervised Learning (SSL): We train the model based on a labeled dataset and an unlabeled dataset,
    \item Fully Supervised Learning (FSL): We train the model with all of the data in both datasets and their true labels.
\end{itemize}

\subsection{Divergence and Entropy}
In this section, we introduce different $f$-divergences and $\alpha$-R\'enyi divergence.

\textbf{$f$-divergence:} The $f$-divergence \cite{polyanskiy2022information}  between two discrete distributions, $P=\{ p_i\}_{i=1}^k$, and $Q=\{q_i\}_{i=1}^k$, is defined as,
\begin{align}
  D_f(P\|Q):=\sum_{i=1}^k q_i f\left(\frac{p_i}{q_i}\right),
\end{align}
where $f:(0,\infty)\rightarrow \mathbb{R}$ is a convex generator function with $f(1)=0$. Note that $D_f(P\|Q)=0$, if $P=Q$. We can also define the $f$-entropy, for discrete distribution $P$ as,
\begin{equation}\label{eq: f-entropy}
     H_f(P)= -D_f(P \| \mathrm{Unif}(k)),
\end{equation}
where $f(\cdot)$ is the same generator function for $f$-divergence. For example, for $f(t)=t\log(t)$, we have $\mathrm{KL}$-divergence, and the entropy is equal to the summation of traditional entropy and a constant term,
\begin{equation}
    H_{\mathrm{KL}}(P)=h_{\mathrm{KL}}(P)-\log(k),
\end{equation}
where $h_{\mathrm{KL}}(P)=-\sum_{i=1}^k P_i\log(P_i)$.

\textbf{$\alpha$-R\'enyi divergence:} The $\alpha$-R\'enyi divergence between $P$ and $Q$ is defined,
\begin{equation}
    D_{\alpha}(P\|Q):=\frac{1}{\alpha-1}\log\Big(\sum_{i=1}^k p_i^\alpha q_i^{1-\alpha}\Big), \quad \alpha\geq 0.
\end{equation}
Similarly, we can define the $\alpha$-R\'enyi entropy of distribution $P$ as follows,
   \begin{equation}
    H_{\alpha}(P):=-D_{\alpha}(P\|\mathrm{Unif}(k)), \quad \alpha\geq 0.
\end{equation} 
Note that our definition of $\alpha$-R\'enyi entropy coincides with the traditional $\alpha$-R\'enyi entropy definition in \cite{van2014renyi},
\begin{equation}\label{eq: alpha entropy}
    H_{\alpha}(P)=h_{\alpha}(P)-\log(k), \quad \alpha\geq 0,
\end{equation} 
where $h_{\alpha}(P):=1/(1-\alpha)\log\big(\sum_{i=1}^k p_i^\alpha\big)$ is traditional $\alpha$-R\'enyi entropy. 
For ease of notation, we define the general divergence and D-entropy as $D(P\|Q)$ and $H_{D}(P)$, where it can be $f$-divergence and $f$-entropy or $\alpha$-R\'enyi divergence and $\alpha$-entropy, respectively.
\subsection{Soft-label And Hard-label}

Our study uses two distinct label types: hard-label and soft-label. In the case of a hard-label, the distribution over the label set is such that $\hat{P}(Y=y_i|X_i)=1$, indicating a certainty that the label is $y_i$, while $\hat{P}(Y=y_j|X_i)=0$ for all $y_j\in\mathcal{Y}$ not equal to $y_i$. Conversely, in the soft-label scenario, we have $\hat{P}(Y=y_j|X_i)\geq 0$ for all labels $y_j$, and $\sum_{j=1}^k \hat{P}(Y=y_j|X_i)=1$. It is worth noting that for labeled datasets, we employ hard-labels. However, for unlabeled datasets, we have the flexibility to adopt either hard-label or soft-label.

\section{Divergence-based empirical risk}
In this section, we introduce divergence-based empirical risk (DER) inspired by divergence, e.g., $f$-divergence and $\alpha$-R\'enyi divergence. All the proofs details are provided in the Appendix.

\subsection{DER For SL Application}
For SL applications, we denote the empirical distribution over the label set for all labeled features by
\begin{equation}
    \widehat{P} (\mbYl|\mbXl):=\Big \{ \frac{1}{n} \hat{P}(\mbYl|X_i^l)\Big\}_{i=1}^{n},
\end{equation}
where $\hat{P}(\mbYl|X_i^l)$ is the empirical true label distribution. Similarly, the estimated conditional distribution of given features by the model with parameters $\theta$ is,
\begin{equation}
    P_{\theta}(\mbY|\mbXl):=\Big\{\frac{1}{n} P_{\theta}(\mbY|X_i^l)\Big\}_{i=1}^{n}. \end{equation}
Note that both $P_{\theta}(\mbY|\mbXl)$ and $\widehat{P}(\mbYl|\mbXl)$ are joint probability distributions over $\mathcal{Y}\times \mbXl$ set, i.e., 
\[\frac{1}{n} \sum_{i=1}^n\sum_{j=1}^k \hat{P}(y_j|X_i^l)=1, \quad \frac{1}{n} \sum_{i=1}^n \sum_{j=1}^k  P_{\theta}(y_j|X_i^l)=1.\]

For the labeled dataset, we consider the hard-label as an empirical distribution over the label set. In particular, for all $X_i^l\in\mbXl$, we can assert that $P(y_i^l|X_i^l)=1$, and for any other label $y_j$ where $y_j\neq y_i^l$, we can assert that $P(y_j|X_i^l)=0$.

The main goal of supervised learning is to learn a model that can predict the true labels of the training data, i.e., $P_{\theta}(\mbY|\mbXl)=\widehat{P}(\mbYl|\mbXl)$. For this purpose, we can consider $D\Big(  \widehat{P}(\mbYl|\mbXl)\|P_{\theta}(\mbY|\mbXl)\Big)$, where the divergence is zero if we have $P_{\theta}(\mbY|\mbXl)=\widehat{P}(\mbYl|\mbXl)$. Therefore, we can define the empirical risk inspired by $f$-divergence or $\alpha$-R\'enyi-divergence between the distributions $\widehat{P}(\mbYl|\mbXl)$ and $P_{\theta}(\mbY|\mbXl)$ as DER, 
\begin{equation}\label{eq: risk alpha-div}
\begin{split}
\hat{R}_{D}(\theta,\mbZl)&:=D\Big(  \widehat{P}(\mbYl|\mbXl)\|P_{\theta}(\mbY|\mbXl)\Big),
\end{split} 
\end{equation}
where $D\in\{D_f,D_\alpha\}$. The true risk is defined as 
\begin{equation*}
   \begin{split}
R_{D}(\theta,\mbZl)&:=\mathbb{E}_{\mbYl,\mbXl}\Big[D\big(  \widehat{P}(\mbYl|\mbXl)\|P_{\theta}(\mbY|\mbXl)\big)\Big],
\end{split}  
\end{equation*}
where we consider the expectation of DER with respect to the distribution of the training dataset.
 In Table~\ref{tab:ferm new}, we provide different DERs based on different $f$-divergences and $\alpha$-R\'enyi-divergence, which satisfy Assumption~\ref{Ass: f-div case}.

Comparing DERs to each other is possible since they are based on divergences. For instance, the $\alpha$-ERM increases as $\alpha$ increases. We also have,
\begin{equation*}
    \lim_{\alpha\rightarrow 1} \hat{R}_{\alpha}(\theta,\mbZl) = \hat{R}_{\mathrm{KL}}(\theta,\mbZl).
\end{equation*}
We can provide the following examples of the inequalities between other DERs,
\begin{itemize}
    \item From \cite{topsoe2000some}, we have $\hat{R}_{\mathrm{KL}}(\theta,\mbZl)\leq \hat{R}_{\chi^2}(\theta,\mbZl)$,
    \item From Pinsker inequality~\cite{canonne2022short}, we have $2\hat{R}_{\mathrm{TV}}^2(\theta,\mbZl)\leq \hat{R}_{\mathrm{KL}}(\theta,\mbZl)$,
    \item From \cite{topsoe2000some}, we have $\hat{R}_{\mathrm{JS}}(\theta,\mbZl)\leq 2\log(2)\hat{R}_{\mathrm{LC}}(\theta,\mbZl)$.
\end{itemize}

Note that some of the DERs are bounded, e.g., $\hat{R}_{\mathrm{TV}}(\theta,\mbZl)\leq 1$, $\hat{R}_{\mathrm{JS}}(\theta,\mbZl)\leq 2\log(2)$, and $\hat{R}_{\mathrm{LC}}(\theta,\mbZl)\leq 1$. 

In addition, $\hat{R}_{\alpha}(\theta,\mbZl)$ is similar to tilted empirical risk~\cite{li2021tilted} by considering the cross-entropy loss in the tilted empirical risk minimization framework. However, our definition is inspired by $\alpha$-R\'enyi divergence. In contrast to~\cite{li2021tilted}, our $\hat{R}_{\alpha}(\theta,\mbZl)$ can be applied to soft-label scenarios. Some of DERs are equivalent to some well-known loss functions. (e.g., $\hat{R}_{\mathrm{TV}}(\theta,\mbZl)$ is equivalent to empirical risk based on mean-absolute-error loss function~\cite{ghosh2017robust} and $\hat{R}_{\mathrm{KL}}(\theta,\mbZl)$ is equivalent to empirical risk based on the cross-entropy loss function.)

\begin{remark}[Comparison with \cite{wei2020optimizing}]
In \cite{wei2020optimizing}, the authors propose a loss function that optimizes the f-mutual information between the true labels and the model predictions. They achieve this by using Fenchel's convex duality for f-divergences to maximize the f-mutual information. However, it is important to note that Fenchel's convex duality framework provides a lower bound on the f-mutual information, and the optimization process focuses on maximizing this lower bound. On the other hand, our framework takes a different approach by minimizing the f-divergence between the empirical distributions of model predictions and true labels. Additionally, our approach can easily accommodate semi-supervised learning and the concept of soft labels.
\end{remark}

\begin{table}[t]
    \centering
     \caption{\small DERs for SL applications, including KL divergence, Total variation distance (TV-distance), $\chi^2$-divergence, Power-divergence (P-divergence), Jensen-Shannon divergence (JS-divergence), Le Cam distance (LC-distance), and $\alpha$-R\'enyi divergence. ``N/A'' means not applicable. We have $P_{i}:=P_\theta(y_i^l|X_i^l)$. \normalsize }
    \small
    \setlength{\tabcolsep}{4pt} 
   \resizebox{\columnwidth}{!}{  \begin{tabular}{cccc}
    \toprule
    \textbf{Divergence} & \textbf{Generator $f(t)$} & \textbf{Definition} & \textbf{DER} \\
    \midrule
    KL-divergence & $t\log(t)$ & $\hat{R}_{\mathrm{KL}}(\theta,\mathbf{Z}^l)$ & $\frac{-1}{n} \sum_{i=1}^n \log(P_i)$ \\
    \midrule
    TV-distance & $\frac{1}{2}|t-1|$ & $\hat{R}_{\mathrm{TV}}(\theta,\mathbf{Z}^l)$ & $\frac{-1}{n} \sum_{i=1}^n (1-P_i)$ \\
    \midrule
    $\chi^2$-divergence & $(1-t)^2$ & $\hat{R}_{\chi^2}(\theta,\mathbf{Z}^l)$ & $\frac{1}{n} \sum_{i=1}^n (P_i^{-1}-1)$ \\
    \midrule
    P-divergence & $t^p-1$ & $\hat{R}_{P}(\theta,\mathbf{Z}^l)$ & $\frac{1}{n} \sum_{i=1}^n (P_i^{-p+1}-1)$ \\
    \midrule
    JS-divergence & $t\log\left(\frac{2t}{1+t}\right)+\log\left(\frac{2}{1+t}\right)$ & $\hat{R}_{\mathrm{JS}}(\theta,\mathbf{Z}^l)$ & $\frac{1}{n} \sum_{i=1}^n \left(P_i\log(P_i) - (P_i+1)\log(P_i+1)\right)$ \\
    \midrule
    LC-distance & $\frac{1-t}{2(1+t)}$ & $\hat{R}_{\mathrm{LC}}(\theta,\mathbf{Z}^l)$ & $\frac{1}{2n} \sum_{i=1}^n \left(1-P_i\right)\left(1-\frac{P_i}{1+P_i}\right) + 2\log(2)$ \\
    \midrule
    $\alpha$-R\'enyi divergence, $\alpha \geq 0$ & N/A & $\hat{R}_{\alpha}(\theta,\mathbf{Z}^l)$ & $\frac{1}{\alpha-1} \log\left(\frac{1}{n} \sum_{i=1}^n P_i^{1-\alpha}\right)$ \\
    \bottomrule
    \end{tabular}}
    \label{tab:ferm new}
\end{table}
\vspace{-1em}
\subsection{DER For SSL Application}
In SSL applications, we focus on self-training approaches, which include methods such as pseudo-labeling and entropy minimization. 
\subsubsection{Pseudo-labeling}
 In this scenario, we assign a pseudo-label to each unlabeled feature through a pseudo-labeling process. We define the pseudo-labeled dataset as $\hat{\mathbf{Z}}:=\{\hat{Y}^j,X_j^u\}_{j=1}^m$, where $\hat{Y}^j$ is the pseudo-labeled assigned to unlabeled data sample. Therefore, we define $\hat{P}(\mbhY|X_j^u)$ as empirical distribution\footnote{The empirical pseudo-label distribution can be either empirical hard pseudo-label or empirical soft pseudo-label distributions.} over unlabeled dataset inspired by pseudo-label generation process for unlabeled feature $X_j^u$. To apply our DER approach in this setup, we define a convex combination of the empirical distribution over label set for all labeled and unlabeled datasets by
 \small
\begin{equation*}\label{Eq: hat dist gamma}
\begin{split}
     &\widehat{P}(\mbYl,\mbhY|\mbXlu):=\\
     &\quad\Big\{\Big\{\frac{\beta}{n} \hat{P}(\mbYl|X_i^l)\Big\}_{i=1}^{n},\Big\{\frac{(1-\beta)}{m} \hat{P}(\mbhY|X_j^u)\Big\}_{j=1}^m\Big\},
\end{split}
\end{equation*}
\normalsize
where $\beta\in[0,1]$.
Similarly, the estimated conditional distribution as a joint distribution over the set $\mbYl\times\mbXlu$
\small
\begin{equation*}\label{Eq: theta dist gamma}
\begin{split}
     &P_{\theta}(\mbY|\mbXlu):=\\&\quad\Big\{\Big\{\frac{\beta}{n} P_{\theta}(\mbY|X_i^l)\Big\}_{i=1}^{n},\Big\{\frac{(1-\beta)}{m} P_{\theta}(\mbY|X_j^u)\Big\}_{j=1}^m\Big\}.
\end{split}
   \end{equation*}
   \normalsize
Note that both $P_{\theta}(\mbY|\mbXlu)$ and $\widehat{P}(\mbYl,\mbhY|\mbXlu)$ are joint probability distributions over $\mathcal{Y} \times \mbXlu$. Similar to \eqref{eq: risk alpha-div}, we can define the DER for the SSL application based on $f$-divergence or $\alpha$-R\'enyi divergence respectively,
\begin{equation*}
    \begin{split}
        \hat{R}_{D}(\theta,\mbZl,\mbZup)&=D\Big(  \widehat{P}(\mbYl,\mbhY|\mbXlu)\|P_{\theta}(\mbY|\mbXlu)\Big),
\end{split}
\end{equation*}
where $D\in\{D_f,D_\alpha\}$. It is worth noting that adjusting the value of $\beta$ allows us to modify the nature of our problem. For instance, setting $\beta$ to 1 creates a supervised learning scenario where the model is trained on labeled data, while $\beta$ being 0 indicates an unsupervised learning scenario. For semi-supervised learning applications, a popular choice is $\beta = \frac{n}{n+m}$. 

Furthermore, certain divergences can function as metrics on the probability distribution space. This characteristic can be utilized to determine the upper limit of the effectiveness of the pseudo-label (or soft-label) approach.

We define $P_t(\mbYl,\mathbf{Y}_t^u|\mbXlu)$ as the true empirical distribution for all labeled and unlabeledd samples, where $\mathbf{Y}_t^u$ is the true labels for the unlabeled samples.

\begin{theorem}\label{thm: main}
Suppose that there exists an increasing function $G:[0,\infty)\to[0,\infty)$ where for a generator function, $f(t)$, $G(D_f(.\|.))$ is a metric on the space of probability distributions. Then, the following holds,
\small
\begin{align*}
&G\Big(\hat{R}_D^{\mathrm{FSL}}(\theta^\star,\mbZl,\mbXu,\mathbf{Y}_t^u)\Big)\\
&\quad\leq G\Big(D_f\Big(P_t(\mbYl,\mathbf{Y}_t^u|\mbXlu)\|\widehat{P}(\mbYl,\mbhY|\mbXlu)\Big)\Big)\\
&\qquad+G\Big(\hat{R}_{D}(\theta^\star,\mbZl,\mbZup)\Big),
\end{align*}
\normalsize
where  $\hat{R}_D^{\mathrm{FSL}}(\theta,\mbZl,\mbXu,\mathbf{Y}_t^u) \:= D_{f}\left( P_t\|P_{\theta^\star} \right),$
is the empirical risk of the FSL scenario, $P_t=P_t(\mbYl,\mathbf{Y}_t^u|\mbXlu)$, $P_{\theta^\star}(\mbY,\mbXlu)$ and $\theta^\star\in \arg \min_{\theta\in\Theta} \hat{R}_{D}(\theta,\mbZl,\mbZup).$
\end{theorem}
It is worth noting that the minimizer of the DER in SSL under some conditions is also a minimizer of the DER in the FSL scenario. We can also derive an upper bound on true risk under the FSL scenario for some generating functions. The following $f$-divergences satisfy the conditions in Theorem~\ref{thm: main},
 \begin{itemize}
     \item Total variation distance for $G(t)=t$,
     \item Le-Cam distance for $G(t)=\sqrt{t}$,~\cite{endres2003new},
     \item Jensen-Shannon divergence for $G(t)=\sqrt{t}$,~\cite{endres2003new}.
 \end{itemize}
 
\subsubsection{Entropy Minimization}
Building upon the ideas presented in \cite{grandvalet2005semi}, we study the concept of D-entropy. In this approach, we compute D-entropy as a regularization term over the distribution of predicted labels, denoted as $P_{\theta}(\mbY|\mbXu)$, for the unlabeled dataset. It's worth noting that the minimization of D-entropy can be interpreted as the maximization of $D(P_{\theta}(\mbY|\mbXu)|\unif)$. Essentially, this means we are actively seeking predicted labels for each unlabeled feature with the maximum dissimilarity with the uniform distribution in terms of $f$-divergence or $\alpha$-R\'enyi divergence. However, the minimization of D--entropy can cause the system to predict the same class for each data sample. 

To avoid the prediction of one specific class for each unlabeled feature, \cite{tanaka2018joint} and \cite{arazo2020pseudolabeling} proposed to use a KL divergence between the mean distribution of Softmax outputs for all unlabeled data samples, i.e.,
$\Bar{P}_\theta(\mbYl|\mbXu):=\frac{1}{m}\sum_{j=1}^m P_{\theta}(\mbY|X_j^u),$
and the uniform distribution. In a similar approach, we propose to minimize the divergence, i.e., $f$-divergence or $\alpha$-R\'enyi divergence, between $\Bar{P}_\theta(\mbYl|\mbXu)$ and uniform distribution. Minimizing this divergence would help the system to predict uniform distribution over all classes. Note that, by the Law of Large Numbers~\cite{hsu1947complete}, if the number of unlabeled data samples goes to infinity $m\rightarrow \infty$, then we have, $\Bar{P}_\theta(\mbYl|\mbXu)\rightarrow \Bar{P}_\theta(\mbYl),$ where $\Bar{P}_\theta(\mbYl)$ is the distribution over all classes that is induced by the algorithm. If we have the balance assumption for all classes, then we expect that $\Bar{P}_\theta(\mbYl)$ would be uniform. Therefore, this regularization can also help in the case when we have an imbalanced number of data samples from classes during the pseudo-labeling process. In particular, after pseudo-labeling (with soft-label or hard-label), we can expect an imbalanced pseudo-labeled dataset. Our final regularized risk for entropy minimization would be,
\begin{align}\label{SSL_LOSS}\nonumber
&\widehat{R}_{\tilde{D}}(\theta,\mbZl,\mbXu,\lambda):=\hat{R}_{\tilde{D}}(\theta,\mbZl) +\lambda_h H_{D}(P_{\theta}(\mbY|\mbXu))\\&\quad+ \lambda_u D(\Bar{P}_\theta(\mbYl|\mbXu)\|\mathrm{Unif}(k)),
\end{align}
Different D-entropy functions are introduced in Appendix~\ref{App: other DERS}. 
\subsubsection{Robustness}\label{sec: robust}
From Corollary~\ref{Cor: main true risk} in Appendix, if SSL scenario's cost, i.e., $D_f\Big(P_t(\mbYl,\mathbf{Y}_t^u|\mbXlu)\|\widehat{P}(\mbYl,\mbhY|\mbXlu)\Big)$, is bounded, then we'll have a notion of robustness with respect to changes in $P_t(\mbYl,\mathbf{Y}_t^u|\mbXlu)$. For example, in the pseudo-labeling process, if the pseudo-label is not equal to the true label of an unlabeled data sample, then we can have noisy pseudo-labels, which increase the SSL scenario's cost. The same holds for soft-label scenarios in the entropy minimization approach. Note that as the total variation distance, Le-Cam distance and the Jensen-Shannon divergence are bounded; therefore, they have a bounded SSL scenario's cost and are robust with respect to pseudo-labeling process.

\section{Algorithms}\label{app: algorithm}
\textbf{DP-SSL:} We propose a divergence-based pseudo-labeling SSL (DP-SSL) algorithm in Algorithm~\ref{Alg: DP-SSL}. In this algorithm, we first generate pseudo-labels for unlabeled data samples in an iterative manner based on an uncertainty-aware process. Let us define $Q(j):=\max_{i\in[k]} P_{\theta}(y_i|X_j^u)$ where $q:=\arg\max_{i\in[k]} P_{\theta}(y_i|X_j^u)$, then we have,
\begin{align}\label{eq: PL gen}
&\hat{\mathrm{Y}}_q^j:=\mathds{1}\big[ Q(j)\geq \tau_p\big]\mathds{1}\big[U( Q(j))\leq \kappa_p\big],
\end{align}
where the function $U:[0,1]\to[0,1]$ estimates the uncertainty of label for a given feature as proposed in \cite{rizve2021in}. If $\hat{\mathrm{Y}}_q^j=0$, then the unlabeled sample would be neglected to reduce the confirmation bias incurred by pseudo-labeling. Otherwise, we select the hard-label for the $q$-th class. Note that the constants $\tau_p$ and $\kappa_p$ are the estimated uncertainty and conditional probability thresholds, respectively. The selection of $\tau_p$ would help us to select the most certain predictions for unlabeled data samples. In addition, increasing the $\tau_p$ would reduce the number of unlabeled samples that can be utilized in the training process. It is worth mentioning that unlabeled data are not included in the first iteration. Therefore, the model derived in the first iteration (Warm-up) is utilized to generate a pseudo-label based on \eqref{eq: PL gen} in the next iteration. After each iteration of the pseudo-labeling process, we balance the set of pseudo-labeled dataset. For this purpose, we under-sample the pseudo-labeled dataset, based on the data samples from the minority class.

 \RestyleAlgo{ruled}
\SetKwComment{Comment}{/* }{ */}
\begin{algorithm}[htb!]
 \DontPrintSemicolon
 \KwData{ $\mbZl=\{(X_i^l,Y_i^l)\}_{i=1}^n$ sampled from $P_{XY}$, $\mbXu=\{X_j^u\}_{j=1}^m$ sampled from $P_X$, hyper-parameters $\beta$, $\tau_p$, $\kappa_p$, $\hat{R}_{D}(\theta,\mbZl)$, and $\hat{R}_{D}(\theta,\ \mbZl \cup \mbZup)$, the $P_\theta$ model based on a divergence, Iteration index by $t_g$ and max Iterations $I$}
  
 \KwResult{A trained neural network with parameter $\theta$ and output of softmax $P_{\theta}$ which minimizes the DER}

$t_g \gets 1$\\
Train model (Warm-Up) $P_\theta$ with SGD based on  $\hat{R}_{D}(\theta,\ \mbZl)$ \\
   \While{$t_g$  $\leq I$}
   {
   1. Select pseudo-labels based on all unlabeled data samples $\mbXu$ based on
    \begin{align*}&\hat{\mathrm{Y}}_q^j=\mathds{1}\big[ Q(j)\geq \tau_p\big]\mathds{1}\big[U( Q(j))\leq \kappa_p\big], \end{align*}\\
    2. $\forall j\in[m]$, if $\hat{\mathrm{Y}}_q^j>0$, then $\mbZup \gets \{(\widehat{X}_j^u,\hat{Y}_q^j)\cup \mbZup\}$\\
    3. Initial your model $P_\theta$\\
    4. $\mbZup\gets \text{Balance}(\mbZup)$\\
    5. Train your model $P_\theta$ with SGD based on 
    $\hat{R}_{D}(\theta, \mbZl \cup \mbZup)$
    \\
    6. $t_g \gets t_g + 1$
   }
\caption{DP-SSL Algorithm}
\label{Alg: DP-SSL}
\end{algorithm}

\textbf{DEM-SSL:} Motivated by the concept of entropy minimization, we introduce a novel approach, Divergence-based entropy minimization Semi-Supervised Learning (DEM-SSL), in this paper. In developing this algorithm, we build upon the techniques presented in \cite{rizve2021in}, incorporating D-entropy minimization. In each iteration of the algorithm, we adopt the previous predictions of unlabeled data samples as soft-labels for these unlabeled data samples. Our objective is to minimize the DER with respect to the true labels for labeled features and the soft-labels assigned to unlabeled data samples. As discussed before, we introduce the minimization of D-entropy and the divergence term $D(\Bar{P}_\theta(\mbYl|\mbXu)|\mathrm{Unif}(k))$ as regularization terms. The utilization of soft-labels for unlabeled data samples serves to reduce confirmation bias, enhancing the effectiveness of our approach.

\section{Experiments And Discussion}
Anonymized code is provided at \href{https://anonymous.4open.science/r/Robust_DEM_SSLv_1}{GitHub link}.

\textbf{DER:} We conduct the experiments for $\mathrm{KL}$-ERM, $\mathrm{JS}$-ERM, $\alpha$-ERM, $\mathrm{P}$-ERM, and $\chi^2$-ERM. As the accuracies of $\mathrm{TV}$-ERM
and $\mathrm{LC}$-ERM are inferior in comparison with other divergences and their slower convergence in training, we have chosen not to present the results for these particular ERMs. For $\mathrm{TV}$-ERM (a.k.a. mean-absolute error), the same phenomena is also observed by \cite{zhang2018generalized}.

 \begin{table}[htb!]
    \caption{\small Comparison of DP-SSL with uncertainty (DP-SSL/WU), DP-SSL without uncertainty (DP-SSL/WOU) and SL algorithms for CIFAR-100 ($n=400$, $m=49600$) and LETTER ($n=104$, $m=17896$) datasets with assuming $\tau_p=0.7$ in DP-SSL algorithm and $\kappa_p=0.005$ for DP-SSL/WU. \normalsize}
    \centering
    \small 
    \centering
 \resizebox{\columnwidth}{!}{ \begin{tabular}{>{\centering\arraybackslash}m{3cm} *{9}{>{\centering\arraybackslash}m{2cm}}}
    \toprule
    \multirow{3}{*}{\textbf{DER}} & \multicolumn{4}{c}{\textbf{LETTER}} & \multicolumn{4}{c}{\textbf{CIFAR-100}} \\
    \cmidrule(lr){2-5} \cmidrule(lr){6-9}
    & \textbf{SL} & \makecell{\textbf{DP-SSL}\\\textbf{/WU}} & \makecell{\textbf{DP-SSL}\\\textbf{/WOU}} & \textbf{FSL} & \textbf{SL} & \makecell{\textbf{DP-SSL}\\\textbf{/WU}} & \makecell{\textbf{DP-SSL}\\\textbf{/WOU}} & \textbf{FSL} \\
    \midrule
    KL & \( {38.77 \pm 1.24} \) & \( {61.57 \pm 0.42} \) & \( {61.72 \pm 0.35} \) & \(61.90 \pm0.67\) &\( {13.89 \pm 0.94} \) & \( \bm{76.38 \pm 0.21} \) & \( \bm{75.52 \pm 1.36} \) & \(\bm{75.24\pm0.10}\)\\
    \( \chi^2 \) & \( {38.25 \pm 0.61} \) & \(56.52\pm0.49\) & \( {56.80 \pm 0.71} \) & \(56.32\pm1.38\) &\( {8.00 \pm 1.06} \) & \( 71.97\pm0.27 \) & \( 71.99 \pm 0.24 \) & \(72.33\pm0.10\)\\
    Pow, ($p=1.2$) & \( {37.13 \pm 0.87} \) & \( 58.88	\pm 0.85 \) & \( {58.75 \pm 0.87} \) &\(59.33\pm0.45\) &\( {13.39 \pm 1.18} \) & \( 75.43 \pm 0.39 \) & \( 75.28 \pm 1.47 \) & \(74.4\pm0.30\)\\
    JS & \( {35.58 \pm 1.89} \) & \( \bm{61.92 \pm 0.73} \) & \( \bm{62.5 \pm 0.90} \) &\(\bm{63.20 \pm 0.58}\) & \( {7.11 \pm 1.06} \) & \( 68.59\pm 0.30 \) & \( 71.89 \pm 0.23 \) & \(71.25	\pm0.60\)\\
    $\alpha$-R\'enyi, ($\alpha=0.6$) & \( \bm{40.01 \pm 0.19} \) & \( 61.30 \pm 0.13 \) & \( 62.0 \pm	0.29 \) & \(61.88 \pm	0.70 \)& \( \bm{13.93 \pm 0.15} \) &  \( 73.15 \pm 0.93 \) &\( {71.01 \pm 0.89} \)  & \(73.66 \pm	0.74\)\\
    \bottomrule
    \end{tabular}}

    \label{table:UPS1}
\end{table}
\vspace{-0.25em}

\textbf{Results and Discussion:} In Table~\ref{table:UPS1}, we conducted experiments involving the DP-SSL algorithm and compared its accuracy with both SL and FSL scenarios. In the case of the DP-SSL algorithm, we set $\tau_p=0.7$.
Furthermore, we explored the impact of the uncertainty term in equation \eqref{eq: PL gen} through two scenarios: one with uncertainty ($\kappa=0.005$) and one without uncertainty. It is noteworthy that, across all DP-SSL algorithms utilizing various divergences, the consideration of uncertainty led to an accuracy improvement of less than $1\%$ in both datasets in many cases. For $\mathrm{JS}$-ERM, we can observe that we have a better accuracy without uncertainty in comparison with uncertainty. For the CIFAR-100 dataset, the $\mathrm{KL}$-ERM achieves the highest accuracy at $76.38\pm0.21$, outperforming other DERs. Among the DERs, $\mathrm{JS}$-ERM achieved the highest accuracy in the LETTER dataset. It is worth noting that, for the SL scenario, the $\alpha$-R'enyi divergence outperformed other DERs in terms of accuracy. As we choose the unlabeled data samples with high confidence, the accuracy of DP-SSL with uncertainty and without uncertainty for some DERs is better than their performance under the FSL scenario, as shown in Table~\ref{table:UPS1}.

As we decrease $\tau_p$, we assign more pseudo-labels to unlabeled data samples. However, this increase in pseudo-labeled data samples is expected to result in noisier pseudo-labels, where we have mismatches between the pseudo-labels and the true labels of the unlabeled data samples. In Table~\ref{table:UPS2}, we conducted experiments for DP-SSL and DEM-SSL algorithms by considering $\tau_p=0.3$ without uncertainty. The $\mathrm{JS}$-ERM has the best performance among other DERs, which is consistent with our robustness discussion in Section~\ref{sec: robust}. Therefore, for a smaller value of $\tau_p$, $\mathrm{JS}$-ERM is more robust in comparison with other DERs. It is worth mentioning that the accuracy of DP-SSL/WOU under $\mathrm{JS}$-ERM for $\tau_p=0.3$ is improved compared to $\tau_p=0.7$. However, the accuracy of DP-SSL/WOU under $\mathrm{KL}$-ERM would decrease from $75.52\pm 1.36$ ($\tau_p=0.7$) to $67.80\pm 0.75$ ($\tau_p=0.3$). We can also observe that DEM-SSL has better performance than DP-SSL in most cases.

\begin{table}[htbp]
\centering
\caption{\small Accuracy of DP-SSL and DEM-SSL. We consider without uncertainty and $\tau_p=0.3$. For DEM-SSL, we assume $\lambda_u = 0.8$ and $\lambda_h = 0.4$.\normalsize}
\label{your-table-label}
 \resizebox{\columnwidth}{!}{\begin{tabular}{lcccc}
    \toprule
    \multirow{3}{*}{\textbf{DER}} & \multicolumn{2}{c}{\textbf{LETTER}} & \multicolumn{2}{c}{\textbf{CIFAR-100}} \\
    \cmidrule(lr){2-3} \cmidrule(lr){4-5}
    & \textbf{DP-SSL/WOU} & \textbf{DEM-SSL/WOU} & \textbf{DP-SSL/WOU} & \textbf{DEM-SSL/WOU} \\
    \midrule
    KL  & $58.87 \pm 2.13$ & $59.14 \pm 0.65$ & $67.80 \pm 0.75$ & $70.49 \pm 0.51$ \\
    $\chi^2$  & $56.52 \pm 0.67$ & $57.60 \pm 0.93$ & $68.02 \pm 1.06$ & $69.05 \pm 0.48$ \\
    Pow, ($p=1.2$)  & $58.55 \pm 1.04$ & $59.10 \pm 0.93$ & $67.20 \pm 0.34$ & $71.14 \pm 0.46$ \\
    JS  & $\bm{61.67 \pm 0.94}$ & $57.49 \pm 1.29$ & $\bm{72.43 \pm 1.06}$ & $\bm{73.34 \pm 0.50}$ \\
    $\alpha$-R\'enyi, ($\alpha=0.6$)  & $57.95 \pm 1.40$ & $\bm{59.65 \pm 2.04}$ & $70.26 \pm 1.31$ & $70.37 \pm 0.60$ \\
    \bottomrule
\end{tabular}}
\label{table:UPS2}
\end{table}
\vspace{-1em}

\section{Conclusion And Future Works}
We provide novel empirical risk functions and regularizers inspired by $f$-divergence and $\alpha$-R\'enyi divergence for self-training algorithms for semi-supervised learning. Our algorithms can be applied to both pseudo-labeling and entropy-minimization. We also discussed, under some divergences, we can provide an upper bound on DERs and their true risks under a fully labeled scenario. Finally, we observe that under more noisy pseudo-labeled or imbalanced data samples, our empirical risk functions are robust. As future works, our framework can be combined with other methods for semi-supervised learning, e.g., Fixmatch~\cite{sohn2020fixmatch}, MixMatch~\cite{berthelot2019mixmatch}, and Meta pseudo-label~\cite{DBLP:journals/corr/abs-2003-10580}.
\section*{Acknowledgements} 
Gholamali Aminian acknowledges the support of the UKRI Prosperity Partnership Scheme (FAIR) under EPSRC Grant EP/V056883/1 and the Alan Turing Institute. 
\clearpage
\newpage
\bibliographystyle{IEEEtran}
\bibliography{ISIT/Reference}

\clearpage
\newpage
\appendices
\section{Related Works}\label{App: related work}
We provide an overview of relevant works concerning self-training techniques in SSL, as well as other SSL methodologies and robust loss functions for handling label noise.  

\textbf{Self-training and SSL}:  Under different scenarios, it is shown that the pseudo-labeling is effective,  \cite{arazo2020pseudolabeling} and \cite{rizve2021in}. \cite{kou2023how} shows that semi-supervised learning with pseudo-labeling can achieve near-zero test loss under some conditions. The study by \cite{DBLP:journals/corr/abs-2003-10580} introduced meta pseudo-labeling. This method enhanced the accuracy of pseudo-labels by incorporating feedback from the student model. \cite{rizve2021in} proposed confidential-based pseudo-label generation for training a network with unlabeled data. \cite{arazo2020pseudolabeling} suggests soft-labeling with the MixUp method to reduce over-fitting to
model predictions and confirmation bias. \cite{oymak2020statistical} and \cite{article_WREPL} analyzed both theoretical and algorithmic side of self-training. In this work, we propose a more general framework as a combination of self-training methods which outperforms previous self-training algorithms.
\newline\textbf{Confirmation Bias:} \cite{arazo2020pseudolabeling} introduced confirmation bias as over-fitting to incorrect pseudo-labels. \cite{Zhang_2018_CVPR} suggests a Mix-Up method to avoid confirmation bias. \cite{yang2022classaware} introduced "Class-aware Contrastive Semi-Supervised Learning" as a method to improve the quality of the pseudo-labels. \cite{cascante2021curriculum} solution is based on re-initializing the model before every self-training iteration. Our work differs from this line of work as we utilize the soft-labels by using D-entropy minimization in DEM-SSL in order to reduce confirmation bias. The performance of the Gibbs algorithm under the SSL scenario is studied in \cite{he2023does}.

\textbf{Other SSL methods:} Some methods use a combination of consistency regularization and pseudo-labeling. 
MixMatch \cite{berthelot2019mixmatch} computes k augmentations for each unlabeled sample, and one for labeled sample in the batch, then sharpens the average output probability of the model for k augmented data and applies the Mix-Up approach~\cite{zhang2018mixup}. Continuing the idea of MixMatch, \cite{berthelot2020remixmatch} introduced ReMixMatch; this method adds distributional alignment between unlabeled and labeled data, moreover,  augmentation anchoring and utilizing the  output of weakly-augmented data as labels for k strongly-augmented unlabeled data.  \cite{li2020dividemix} established DivideMix proposed a new method for learning with noise based on the Gaussian Mixture Model (GMM) and MixMatch method. 
\cite{sohn2020fixmatch} presents FixMatch, which uses weakly-augmented input model prediction pseudo-label as a label for strongly-augmented input model prediction. This line of research differs from ours as our focus is self-training algorithms despite consistency regularization methods. 

\textbf{Robust loss functions to label-noise:} The pioneering work by \cite{ghosh2017robust} proved that the mean absolute error loss function is robust to symmetric and label-dependent noises. 
On the other hand, the Generalized cross-entropy loss function, which can be reduced to mean absolute error, and the cross-entropy loss function, is proposed by \cite{zhang2018generalized}. Generalized Jensen-Shannon as a loss function robust to label noise is proposed by \cite{englesson2021generalized}. The loss functions based on $f$-divergence between the joint and product of marginal distributions of clean and noisy labels for the label-noise scenario are studied by \cite{wei2020optimizing}. The work by \cite{xu2019ldmi} proposes an information-theoretic loss function using determinant-based mutual information, which is robust to instance-independent label noise. The normalizing technique is applied to some loss functions, e.g., cross-entropy and focal loss, by \cite{ma2020normalized} to make these loss functions robust to label noise. The peer loss functions based on the peer prediction mechanism are studied by \cite{liu2020peer}.
Since the pseudo-labels can mismatch with the true label of an unlabeled data sample, we can model this process as supervised learning with noisy labels. Our work differs from this body of research in the sense that we provide general novel empirical risk functions and regularizers inspired by divergences for the SSL applications. Note that pseudo-labels are a type of input-dependent label noise, and our proposed algorithms based on these empirical risk functions and regularizers are robust to the noise of pseudo-labels.

\section{Theoretical Results and Proofs}\label{app: proofs}

Note that, the DER is not well-defined for all f-divergences. For this purpose, we consider the following assumption.
\begin{assumption}\label{Ass: f-div case}
    The generator function of $f$-divergence satisfies $f(0)<\infty$.
\end{assumption}
\begin{proposition}\label{Prop: f-div}
    Under Assumption~\ref{Ass: f-div case} and assuming hard-label for the labeled dataset, the DER based on $f$-divergence exists, and we have 
    \[0\leq\hat{R}_{D_f}(\theta,\mbZl)<\infty.\]
\end{proposition}
\begin{IEEEproof}[Proof of Proposition~\ref{Prop: f-div}]
    From the definition of DER and $f$-divergence we have,
    \begin{equation}\label{eq: pr 1}
\begin{split}
\hat{R}_{D_f}(\theta,\mbZl)&=D_f\Big(  \widehat{P}(\mbYl|\mbXl)\|P_{\theta}(\mbY|\mbXl)\Big)\\
&=\frac{1}{n}\sum_{i=1}^n\sum_{j=1}^k P_{\theta}(y_j|X_i^l) f\Big(\frac{\widehat{P}(y_j|X_i^l)}{P_{\theta}(y_j|X_i^l)}\Big) ,
\end{split} 
\end{equation}
If we consider a hard-label, then there exists $j\in[k]$ and $i\in[n]$, where $\widehat{P}(y_j|X_i^l)=0$ and $P_{\theta}(y_j|X_i^l)>0$ and we have $P_{\theta}(y_j|X_i^l) f(0)$. Therefore, for $f(0)<\infty$, DER in \eqref{eq: pr 1} is well defined. Otherwise, DER is infinite. 
\end{IEEEproof}
For example, considering reverse KL-divergence with $f(t)=-\log(t)$ and symmetrized KL-divergence with $f(t)=(t-1)\log(t)$, do not satisfy Assumption~\ref{Ass: f-div case} and are infinite if we consider the hard-label for the labeled data samples.

\begin{IEEEproof}[Proof of Theorem~\ref{thm: main}]
    As $G(D_f(.\|.))$ is a metric on the space of probability distribution, then for $P_1, P_2$ and $P_3$ as distributions,
    \begin{equation}\label{eq: metric}
        G(D_f(P_1\|P_3))\leq G(D_f(P_1\|P_2))+G(D_f(P_2\|P_3)),
    \end{equation}
    If we consider,
    \[\begin{split}
    P_1&=P_t(\mbYl,\mathbf{Y}_t^u,\mbXlu),\\
    P_2&=\widehat{P}(\mbYl,\mbhY|\mbXlu),\\
    P_3&=P_{\theta^\star}(\mbY,\mbXlu),
    \end{split}\]
    in \eqref{eq: metric}, the final result holds by considering $\hat{R}_D^{\mathrm{FSL}}(\theta,\mbZl,\mbXu,\mathbf{Y}_t^u)= D_{f}\Big( P_t(\mbYl,\mathbf{Y}_t^u,\mbXlu)\|P_{\theta^\star}(\mbY,\mbXlu)\Big)$ and $\hat{R}_{D}(\theta,\mbZl,\mbhY,\mathbf{X}_m^u)=D_f\Big(  \widehat{P}(\mbYl,\mbhY|\mbXlu)\|P_{\theta}(\mbY|\mbXlu)\Big)$.
\end{IEEEproof}

\begin{corollary}\label{Cor: main true risk}
    Assume that there exists an increasing and concave function $G:[0,\infty)\rightarrow[0,\infty)$ such that $2G(t/2)\leq G(2t)$ and $G(D_f(\cdot\|\cdot))$ is a metric on the space of probability distributions. Then, the following holds,
    \small
    \begin{align*}
         \begin{split}
        &R_D^{\mathrm{FSL}}(\theta^\star_t,\mbZl,\mbXu,\mathbf{Y}_t^u)\leq \\&2\mathbb{E}_{\mbZl,\mbZup,\mathbf{Y}_t^u}\Big[D_f\Big(P_t(\mbYl,\mathbf{Y}_t^u|\mbXlu)\|\widehat{P}(\mbYl,\mbhY|\mbXlu)\Big)\Big]\\&\qquad+2R_{D}(\theta^\star_t,\mbZl,\mbZup),
    \end{split}
    \end{align*}
    \normalsize
   where  $R_D^{\mathrm{FSL}}(\theta,\mbZl,\mbXu,\mathbf{Y}_t^u):= \mathbb{E}_{\mbZl,\mbXu,\mathbf{Y}_t^u}\big[D_{f}\big( P_t\|P_{\theta^\star}\big)\big],$ is the true risk of FSL scenario, $P_t=P_t(\mbYl,\mathbf{Y}_t^u,\mbXlu)$, $P_{\theta^\star}=P_{\theta^\star}(\mbY,\mbXlu)$ , and $\theta^\star_t\in \arg \min_{\theta\in\Theta} R_{D}(\theta,\mbZl,\mbZup).$
    \end{corollary}
\begin{IEEEproof}[Proof of Corollary~\ref{Cor: main true risk}]
    From Theorem~\ref{thm: main}, we have,
     \begin{align}\label{eq: cor 1}
        &G\Big(\hat{R}_D^{\mathrm{FSL}}(\theta^\star,\mbZl,\mbXu,\mathbf{Y}_t^u)\Big)\\\nonumber&\leq G\Big(D_f\Big(P_t(\mbYl,\mathbf{Y}_t^u|\mbXlu)\|\widehat{P}(\mbYl,\mbhY|\mbXlu)\Big)\Big)\\\nonumber&\quad+G\Big(\hat{R}_{D}(\theta^\star,\mbZl,\mbZup)\Big)\\\label{eq: cor 2}
        &\leq 2G\Big(\frac{1}{2}D_f\Big(P_t(\mbYl,\mathbf{Y}_t^u|\mbXlu)\|\widehat{P}(\mbYl,\mbhY|\mbXlu)\Big)\\\nonumber&\quad+\frac{1}{2}\hat{R}_{D}(\theta^\star,\mbZl,\mbZup)\Big)\\\label{eq: cor 3}
        &\leq G\Big(2D_f\Big(P_t(\mbYl,\mathbf{Y}_t^u|\mbXlu)\|\widehat{P}(\mbYl,\mbhY|\mbXlu)\Big)\\\nonumber&\quad+2\hat{R}_{D}(\theta^\star,\mbZl,\mbZup)\Big),
    \end{align}
    where \eqref{eq: cor 1}, \eqref{eq: cor 2} and \eqref{eq: cor 3} follow from Theorem~\ref{thm: main}, concavity of function $G(\cdot)$ and the assumption that $2G(t/2)\leq G(2t)$, respectively. As we have  \begin{align}\label{eq: cor 4}
        &G\Big(\hat{R}_D^{\mathrm{FSL}}(\theta^\star,\mbZl,\mbXu,\mathbf{Y}_t^u)\Big)\\\nonumber&\leq  G\Big(2D_f\Big(P_t(\mbYl,\mathbf{Y}_t^u|\mbXlu)\|\widehat{P}(\mbYl,\mbhY|\mbXlu)\Big)\\\nonumber&\quad+2\hat{R}_{D}(\theta^\star,\mbZl,\mbZup)\Big),
    \end{align} 
   From increasing assumption on function function $G(\cdot)$, we have,
   \begin{align}\label{eq: cor 5}
        &\hat{R}_D^{\mathrm{FSL}}(\theta^\star,\mbZl,\mbXu,\mathbf{Y}_t^u)\\\nonumber&\leq 2D_f\Big(P_t(\mbYl,\mathbf{Y}_t^u|\mbXlu)\|\widehat{P}(\mbYl,\mbhY|\mbXlu)\Big)\\\nonumber&\quad+2\hat{R}_{D}(\theta^\star,\mbZl,\mbZup)\Big).
    \end{align} 
    The final result holds by taking the expectation from both sides of \eqref{eq: cor 5} with respect $\mbZl,\mbZup$ and $\mathbf{Y}_t^u$.
\end{IEEEproof}
Note that, for DERs in Corollary~\ref{Cor: main true risk} ,the term $D_f\big(P_t(\mbYl,\mathbf{Y}_t^u|\mbXlu)\|\widehat{P}(\mbYl,\mbhY|\mbXlu)\big)$ is independent from $\theta$ and can be interpreted as cost of SSL scenario. For example, if the true risk DER under the SSL scenario is zero, then for the same minimizer, $\theta^\star_t$, we can bound the FSL scenario's true risk with the SSL scenario's cost.

The following $f$-divergences satisfy the conditions in Corollary~\ref{Cor: main true risk},
 \begin{itemize}
     \item Total variation distance for $G(t)=t$,
     \item Le-Cam distance for $G(t)=\sqrt{t}$,~\cite{endres2003new},
     \item Jensen-Shannon divergence for $G(t)=\sqrt{t}$,~\cite{endres2003new}.
 \end{itemize}
 
 \begin{remark}[$\mathrm{TV}$-ERM]
  We can provide a tighter upper bound for on the true risk of the FSL scenario under $\mathrm{TV}$-ERM in comparison with Corollary~\ref{Cor: main true risk}, as follows,
     \begin{equation}
         \begin{split}
        &\hat{R}_{\mathrm{TV}}^{\mathrm{FSL}}(\theta^\star,\mbZl,\mbXu,\mathbf{Y}_t^u)\\&\quad\leq \mathrm{TV}\Big(P_t(\mbYl,\mathbf{Y}_t^u|\mbXlu),\widehat{P}(\mbYl,\mbhY|\mbXlu)\Big)\\&\qquad+R_{\mathrm{TV}}(\theta^\star,\mbZl,\mathbf{X}_m^u).
    \end{split}
    \end{equation}
 \end{remark}

\begin{remark}[Comparison with \cite{aminian2022information} and \cite{he2022information}]
    We can also define the ERM for SSL application based on a convex combination of ERM for labeled dataset and unlabeled dataset, as proposed in \cite{aminian2022information} and \cite{he2022information}. However, due to the convexity of $f$-divergences, our DER for SSL application is a lower bound for the proposed setup,
\begin{equation*}
    \begin{split}
        &D\Big(  \widehat{P}(\mbYl,\mbhY|\mbXlu)\|P_{\theta}(\mbY|\mbXlu)\Big)\\&\leq \beta D\Big(  \widehat{P}(\mbYl|\mbXl)\|P_{\theta}(\mbY|\mbXl)\Big)\\&\quad +(1-\beta)D\Big(  \widehat{P}(\mbhY,\mbXu)\|P_{\theta}(\mbY,\mbXu)\Big).
\end{split}
\end{equation*}
\end{remark} 

 \section{Robustness Discussion}
The robustness of our DERs in SSL applications is different from the label-noise approach in \cite{ghosh2017robust}. Our approach can be applied to both soft-label and hard-label. However, robust loss functions to label-noise in \cite{ghosh2017robust} are discussed for hard-labels. Our robustness definition is inspired by f-divergences, which are metric over spaces and satisfy the assumptions in Theorem~\ref{thm: main}. In addition, these $f$-divergences can be applied to all types of noise. However, the results in \cite{ghosh2017robust} are based on symmetric loss function definitions. 
 \section{D-Entropy}\label{App: other DERS}
Different DERs and the corresponding entropy are introduced in Table~\ref{tab:ferm_new2}.

\begin{table*}[tbh]
    \centering
     \caption{DER and D-Entropy for $\alpha$-R\'enyi, as well as metrics like KL divergence, Power divergence, JS divergence, Le Cam, and Total variation distance. We have \( P_{i} := P_\theta(y_i^l|X_i^l) \).}
    \small
     \resizebox{0.6\textwidth}{!}{ \begin{tabular}{>{\centering\arraybackslash}m{1.5in} >{\centering\arraybackslash}m{2in} >{\centering\arraybackslash}m{2in}}
        \toprule
        \textbf{Name/ Generator \( f(t) \)} & \textbf{DER} & \textbf{D-Entropy } \\
        \midrule
        
        KL,\\ $t\log(t)$ &
        $\begin{aligned}[c]
        &\frac{-1}{n} \sum_{i=1}^n \log(P_i)
        \end{aligned}$ &
        $-\log k - \sum_{i=1}^k P_i \log P_i$ \\
        
        \midrule
        TV,\\ $\frac{1}{2}|t-1|$ &
        $\frac{1}{n}\Big(\sum_{i=1}^n (1-P_i)\Big)$ &
        $\frac{-1}{2} \sum_{i=1}^k |P_i - \frac{1}{k}|$ \\
        
        \midrule
        $\chi^2$,\\ $(1-t)^2$ &
        $\frac{1}{n}\Big(\sum_{i=1}^n (P_i^{-1}-1)\Big)$ &
        $-\frac{1}{k} \sum_{i=1}^k (1-kP_i)^2$ \\
        
        \midrule
        Power,\\ $t^p-1$ &
        $\frac{1}{n}\Big(\sum_{i=1}^n (P_i^{-p+1}-1)\Big)$ &
        $1 - k^{p-1} \sum_{i=1}^k P_i^p$ \\
        
        \midrule
        Jensen-Shannon,\\ $t\log\left(\frac{2t}{1+t}\right)+\log\left(\frac{2}{1+t}\right)$ &
        $\begin{aligned}[c]
        \frac{1}{n} \Big( \sum_{i=1}^n P_i\log(P_i) & 
        - (P_i + 1)\log\left(P_i + 1\right) \Big) \\ + &2\log(2) 
        \end{aligned}$ &
        $\begin{aligned}[c]
        &- \sum_{i=1}^k P_i \log \left(1 + \frac{1}{kP_i}\right) \\  &+ \sum_{i=1}^k \frac{1}{k} \log (1 + kP_i) \\  &\qquad-2\log(2)\end{aligned}$
        \\
        \midrule
        Le Cam,\\ $\frac{1-t}{2(1+t)}$ &
        $\frac{1}{2n}\sum_{i=1}^n (1-P_i) \left(1-\frac{P_i}{1+P_i}\right)$ &
        $\sum_{i=1}^k \frac{kP_i - 1}{2k(1+kP_i)}$ \\
        
        \midrule
        $\alpha$-R\'enyi,\\ N/A &
        $\frac{1}{\alpha-1}\log\Big(\frac{1}{n}\sum_{i=1}^n P_i^{1-\alpha} \Big)$ &
        $\log k + \frac{1}{\alpha - 1}\log \sum_{i=1}^k \sum_{j=1}^n P_\theta(y_j|X_i^l)^{\alpha}$ \\
        
        \bottomrule
    \end{tabular}}
   
    \label{tab:ferm_new2}
\end{table*}
\section{Experiment Details}\label{app:experiment}

\textbf{Datasets:} We ran different experiments to validate our proposed algorithms, DEM-SSL and DP-SSL, on two datasets: CIFAR-100 \cite{Krizhevsky2009LearningML} and the Letter \cite{chang2011libsvm} datasets. For the SSL scenario, we have allocated $n=104$ labeled data samples and $m=17896$ unlabeled data samples for the Letter dataset and $n=400$ labeled data samples and $m=49600$ unlabeled data samples for CIFAR-100. We utilized the CNN-13 network 
architecture for CIFAR-100 (\cite{iscen2019label}, \cite{Shi2018TransductiveSD}, \cite{tarvainen2018mean}, \cite{Verma_2022}, \cite{ke2019dual},\cite{berthelot2020remixmatch}, 
\cite{DBLP:journals/corr/abs-1806-05594}) and 2-layer Feedforward neural network inspired by  \cite{DBLP:journals/corr/abs-2112-14869} for letter.

\textbf{Hyper-parameters:} We use a combination of manual and automatic hyper-parameter tuning for the learning rate values and regularization coefficients. For parameter $\beta$, we select $\beta=\frac{n}{n+m}$. We have two hyper-parameters for DP-SSL, i.e., $\tau_p$ and $\kappa_p$. We set $\tau_p\in \{0.3,0.7\} $ and $\kappa_p=0.005$ in \eqref{eq: PL gen}. For DEM-SSL, regularization weights ($\lambda_u , \lambda_h$) inspired by \cite{arazo2020pseudolabeling} and running cross-validation, we selected $\lambda_u = 0.8$ and $\lambda_h = 0.4 $ for DEM-SSL across all DERs. More details are provided in Table~\ref{tab:experiment_setup}.

\begin{table}[htb!]
    \centering
    \caption{Experiment setup details for CIFAR-100 and Letter}
    \label{tab:experiment_setup}
    \begin{tabular}{ccc}
        \toprule
        & CIFAR-100 & Letter \\
        \midrule
        Optimizer & SGD & SGD \\
        Learning rate & 0.03 & 0.03 \\
        Network & CNN-13 & FFNN \\
        Max epochs ($M$) & 512 & 512 \\
        Labeled dataset size ($n$) & 400 & 104 \\
        Unlabeled dataset size ($m$) & 49600 & 17896 \\
        Train/Test size & 50000/10000 & 18000/2000 \\ 
        Batch size & 512 & 512 \\
        Max Iterations ($I$) & 5 & 5 \\
        $\lambda_u$ & 0.8 & 0.8 \\
        $\lambda_h$ & (0.4, 0.04) & (0.4, 0.04) \\
        $\tau_p$ & (0.3,0.7) & (0.3,0.7) \\
        $\kappa_p$ & 0.05 & 0.05 \\
        $\beta$ & 0.992 & 0.994 \\
        \bottomrule
    \end{tabular}
\end{table}


We used 20\%/80\% of CIFAR-100 and 10\%/90\% of Letter datasets for the test/training process. In the FSL scenario, we only train our network with all 80\% of labeled data. The implementation uses the PyTorch framework \cite{paszke2019pytorch}, training was optimized using SGD with nesterov momentum of $0.9$ \cite{botev2016nesterovs}, learning rate of $0.03$, cosine annealing for five iterations and $512$ epoch for each iteration. Experiments are executed on Nvidia Volta V100 GPU with 32 GB VM.

\section{Additional Experiments}
\textbf{No balancing:} As mentioned in DP-SSL and DEM-SSL, after each pseudo-labeling iteration, we balance the pseudo-labeled data samples. In Table~\ref{table:UPS3}, we conducted DP-SSL and DEM-SSL algorithms without balancing (imbalance) in order to show how DP-SSL and DEM-SSL can handle imbalance pseudo-labels in the training stage. Note that in this setup, we set $\tau_p=0.3$ and do not consider uncertainty. We can observe that under the imbalance scenario in pseudo-labeled data samples, the $\chi^2$-ERM has a better performance in comparison with other DERs. For example, the accuracy of $\chi^2$-ERM under balancing and imbalance for $\tau_p=0.3$ and DEM-SSL in CIFAR-100  is $54.17\pm 0.50$ and $50.0\pm 0.48$, respectively.
\begin{table}[htbp]
    \caption{Accuracy of DP-SSL and DEM-SSL under no Balancing. We consider $\tau_p=0.3$ for DP-SSL. For DEM-SSL, we assume $\lambda_u = 0.8$ and $\lambda_h = 0.04$.}
    \centering
    \resizebox{\columnwidth}{!}{ \begin{tabular}{>{\centering\arraybackslash}m{3cm} *{4}{>{\centering\arraybackslash}m{2cm}}}
        \toprule
        \multirow{3}{*}{\textbf{DER}} & \multicolumn{2}{c}{\textbf{LETTER}} & \multicolumn{2}{c}{\textbf{CIFAR-100}} \\
        \cmidrule(lr){2-3} \cmidrule(lr){4-5}
        & \makecell{\textbf{DP-SSL}\\\textbf{/NB\&WOU}} & \makecell{\textbf{DEM-SSL}\\\textbf{/NB\&WOU}} & \makecell{\textbf{DP-SSL}\\\textbf{/NB\&WOU}} & \makecell{\textbf{DEM-SSL}\\\textbf{/NB\&WOU}} \\
        \midrule
        KL  & $45.55 \pm 0.75$ & $52.1 \pm 2.48$ & $19.46 \pm 0.24$ & $35.84 \pm 0.94$ \\
        $\chi^2$ & $\mathbf{53.9 \pm 1.25}$ & $\mathbf{54.17 \pm 0.50}$ & $\mathbf{43.14 \pm 0.47}$ & $\mathbf{50.0\pm0.48}$ \\
        Pow, ($p = 1.2$) & $43.74 \pm	0.56$ & $53.7 \pm 1.09$ & $31.45 \pm 0.11$ & $45.36 \pm 1.18$ \\
        JS  & $39.05 \pm 1.15$ & $41.13 \pm 1.01$ & $7.46 \pm 0.12$ & $46.45 \pm	2.16$ \\
        $\alpha$-R\'enyi, $(\alpha = 0.6)$ & $45.00 \pm	0.70$ & $47.15 \pm	0.74$ & $19.37 \pm	0.10$ & $29.86\pm0.4$ \\
        \bottomrule
    \end{tabular}}
    \label{table:UPS3}
\end{table}
\newpage
\textbf{TV-ERM and LeCam-ERM:}   TV and LeCam results for setup of Table. \ref{table:UPS1} are presented in Table. \ref{table:TV_LeCam}. Due to the lack of space, we did not present them in Table~\ref{table:UPS1}. The poor performance of TV-ERM could be due to the fact that its derivative ($\pm \frac{1}{2}$) is constant.

 \begin{table}[htbp!]

    \caption{Comparison of DP-SSL with uncertainty (DP-SSL/WU), DP-SSL without uncertainty (DP-SSL/WOU) and SL algorithms for CIFAR-100 ($n=400$, $m=49600$) and LETTER ($n=104$, $m=17896$) datasets with assuming $\tau_p=0.7$ in DP-SSL algorithm and $\kappa_p=0.005$ for DP-SSL/WU.}
    \small 
 \resizebox{\columnwidth}{!}{ \begin{tabular}{>{\centering\arraybackslash}m{3cm} *{9}{>{\centering\arraybackslash}m{2cm}}}
    \toprule
    \multirow{3}{*}{\textbf{DER}} & \multicolumn{4}{c}{\textbf{LETTER}} & \multicolumn{4}{c}{\textbf{CIFAR-100}} \\
    \cmidrule(lr){2-5} \cmidrule(lr){6-9}
    & \textbf{SL} & \makecell{\textbf{DP-SSL}\\\textbf{/WU}} & \makecell{\textbf{DP-SSL}\\\textbf{/WOU}} & \textbf{FSL} & \textbf{SL} & \makecell{\textbf{DP-SSL}\\\textbf{/WU}} & \makecell{\textbf{DP-SSL}\\\textbf{/WOU}} & \textbf{FSL} \\
    \midrule
    TV & \( {16.85 \pm 2.84} \) & \( {40.70 \pm	2.00} \) & \( {40.75 \pm 2.94} \) & \( 41.03 \pm 0.65 \) &\( {3.40 \pm 1.94} \) & \( {11.32 \pm	0.71} \) & \( {11.73 \pm 1.34} \) & \({20.64 \pm 1.10}\)\\
    LeCam & \( {22.4 \pm 1.70} \) & \( 57.34	\pm 0.96 \) & \( {58.30 \pm 0.65} \) &\( 59.84 \pm 1.23 \) &\( {12.49 \pm 1.18} \) & \( 55.13 \pm 3.87 \)  & \( 60.12 \pm 1.32 \) & \(63.14	 \pm	2.99\)\\
    \bottomrule
    \end{tabular}}

    \label{table:TV_LeCam}
\end{table}

\end{document}